\documentclass[letterpaper, 10 pt, journal, twoside]{ieeetran}
\usepackage{hyperref}
\usepackage{url}
\usepackage{graphics} 
\usepackage{times} 
\usepackage{booktabs}       
\usepackage{amsfonts}       
\usepackage{nicefrac}       
\usepackage{microtype}      
\usepackage{algorithm}
\usepackage{algorithmic}
\usepackage{mathtools}
\usepackage{graphicx}
\usepackage{cases}
\usepackage{array}
\usepackage{color}
\usepackage{float}
\usepackage{amssymb}
\usepackage{wrapfig}
\usepackage{subfigure}
\usepackage{amsmath}
\usepackage{soul}
\usepackage{amsthm}
\usepackage{graphicx} 
\usepackage{subfigure}
\usepackage{epstopdf}
\usepackage{multirow}
\usepackage{diagbox}
\usepackage{xcolor}
\usepackage{paralist} 
\usepackage{arydshln} 
\theoremstyle{definition}

\theoremstyle{remark}

\begin{document}

\title{ EFEAR-4D: Ego-Velocity Filtering for Efficient and Accurate 4D Radar Odometry}

\author{Xiaoyi Wu$^{1}$, Yushuai Chen$^{1}$, Zhan Li $^{2}$, Ziyang Hong$^{1*}$ and~Liang~Hu$^{1*}$,~\IEEEmembership{Senior Member,~IEEE}
\thanks{Manuscript received: May, 20, 2024; Revised August, 7, 2024; Accepted September, 5, 2024.}
\thanks{This paper was recommended for publication by Editor Sven Behnke upon evaluation of the Associate Editor and Reviewers' comments.} 
\thanks{This work is supported in part by Shenzhen Science and Technology Program (Project No. JCYJ20220818103000001) and Guangdong Provincial Key Laboratory of Intelligent Morphing Mechanisms and Adaptive Robotics (Project No. 2023B1212010005).}
\thanks{$^{1}$ X.~Wu, Y.~Chen, Z.~Hong and L.~Hu are with the Department
of Automation, School of Mechanical Engineering and Automation, Harbin Institute of Technology, Shenzhen, 518055, 
China. $^{2}$ Z.~Li is with Research Institute of Intelligent Control and Systems, Harbin Institute of Technology, Harbin, 150001, China.  *Corresponding authors: {\tt\footnotesize Ziyang Hong (l.hu@hit.edu.cn); Liang Hu (l.hu@hit.edu.cn)}.}
}


\maketitle

\begin{abstract}
Odometry is a crucial component for successfully implementing autonomous navigation, relying on sensors such as cameras, LiDARs and IMUs. However, these sensors may encounter challenges in extreme weather conditions, such as snowfall and fog. The emergence of FMCW radar technology offers the potential for robust perception in adverse conditions. As the latest generation of FWCW radars, the 4D mmWave radar provides point cloud with range, azimuth, elevation, and Doppler velocity information, despite inherent sparsity and noises in the point cloud.  EFEAR-4D exploits Doppler velocity information delicately for robust ego-velocity estimation, resulting in a highly accurate prior guess. EFEAR-4D maintains robustness against point-cloud sparsity and noises across diverse environments through dynamic object removal and effective region-wise feature extraction. Extensive experiments on two publicly available 4D radar datasets demonstrate state-of-the-art reliability and localization accuracy of EFEAR-4D under various conditions. Furthermore, we have collected a dataset following the same route but varying installation heights of the 4D radar, emphasizing the significant impact of radar height on point cloud quality, a crucial consideration for real-world deployments. Our algorithm and dataset are available at \url{https://github.com/CLASS-Lab/EFEAR-4D}.
\end{abstract}

\begin{IEEEkeywords}
SLAM, Autonomous Vehicle Navigation, 4D mmWave radar, Radar Odometry
\end{IEEEkeywords}

\section{Introduction}\label{section:Intro}
\IEEEPARstart{O}{dometry} estimation is a core challenge for autonomous navigation, particularly in GPS-denied environments \cite{adolfsson2022LiDAR,zhuoins20234drvo}. Over the past decades, vision-based and LiDAR-based odometry has been greatly advanced by the progress in sensing technologies, achieving high performance in various real-time applications \cite{shan2020lio}. However, both LiDAR and camera odometry systems still suffer from performance degradation under adverse weather conditions \cite{adolfsson2022LiDAR, zhang20234DRadarSLAM, harlow2024new}. To maintain robust all-weather localization capability, several odometry methods resort to millimeter-wave (mmWave) radars as an alternative. \cite{adolfsson2022LiDAR, zhang20234DRadarSLAM, hong2022radarslam, zhuang20234d, chen2023drio}. Compared with the LiDAR, the radar exhibits a longer wavelength and is minimally affected by adverse weather conditions such as rain, snow, and fog. The ability to detect extended ranges and provide Doppler velocity information constitutes an additional advantage, facilitating the calculation of ego-velocity \cite{kellner2013instantaneous}. 

Compared with its predecessor 2D and 3D mmWave radars, 4D mmWave radars (called 4D radars for shortness thereafter) can provide more information in additional dimensions, that is elevation (3D radars have it as well) and Doppler velocity, but with sparser point cloud due to more compact device size and lower power consumption. These unique pros and cons necessitate developing a tailored and robust odometry solution for 4D radars. However, two challenges that need to be addressed have hindered developing effective 4D radar odometry methods. The first challenge comes from sparsity in point cloud and exclusive radar noises such as multi-path propagation and harmonics \cite{zhuoins20234drvo, zeller2022gaussian}, which make it hard to extract geometric features such as edges and planes. The second challenge comes from the limited field of view (FOV) of 4D radars, typically around $120^\circ$ in the horizontal plane and $40^\circ$ in the vertical plane, which makes reliable data association among frames challenging. 

\begin{figure}[t]
    \centering
    \includegraphics[width=0.48\textwidth]{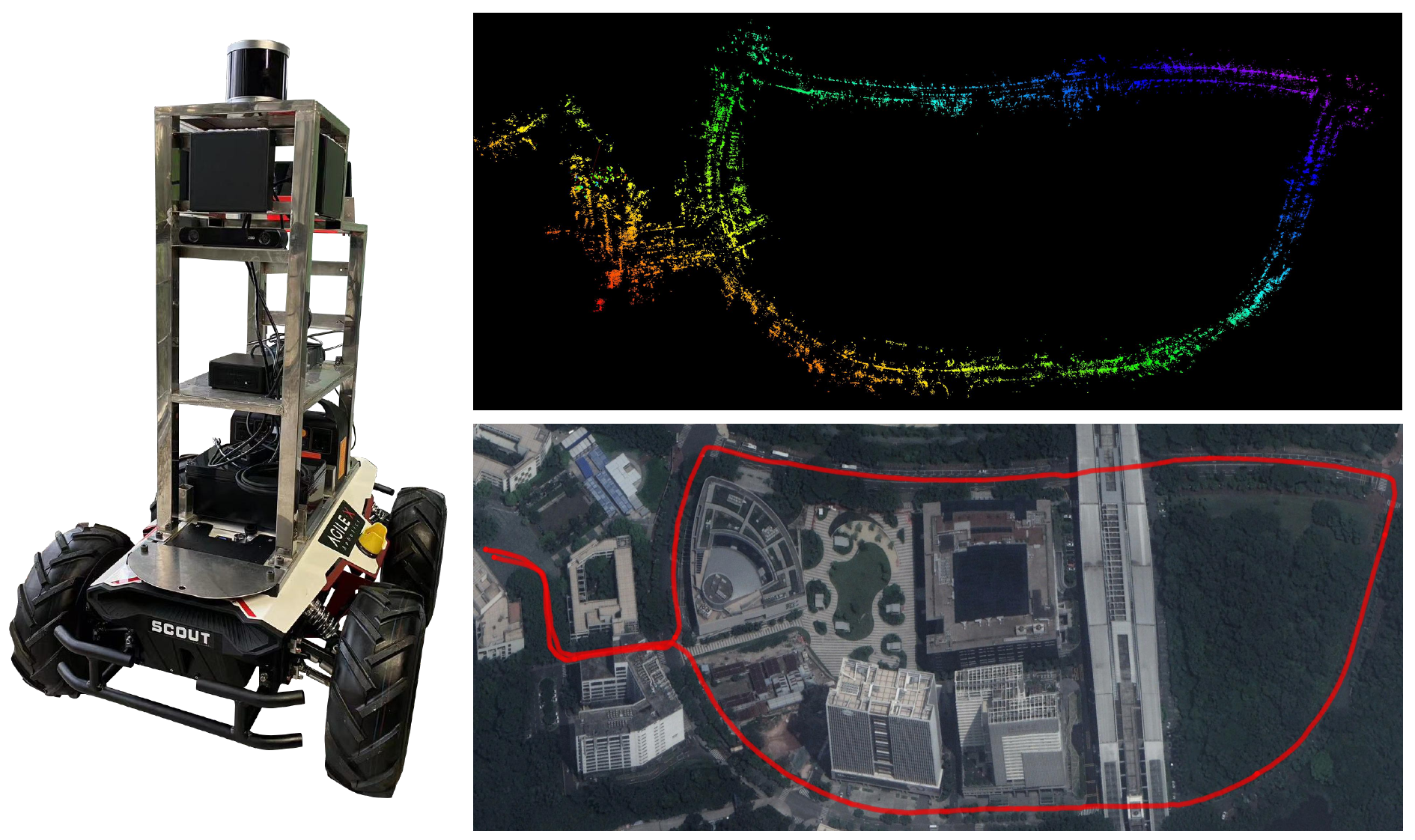}
    \caption{ Our efficient 4D radar odometry and mapping. We test our system on our AGV platform. The mapping result and corresponding ground-truth are depicted on the right.}
    \label{fig1}
\end{figure}
\begin{figure*}
    \centering
\includegraphics[width=0.95\textwidth]{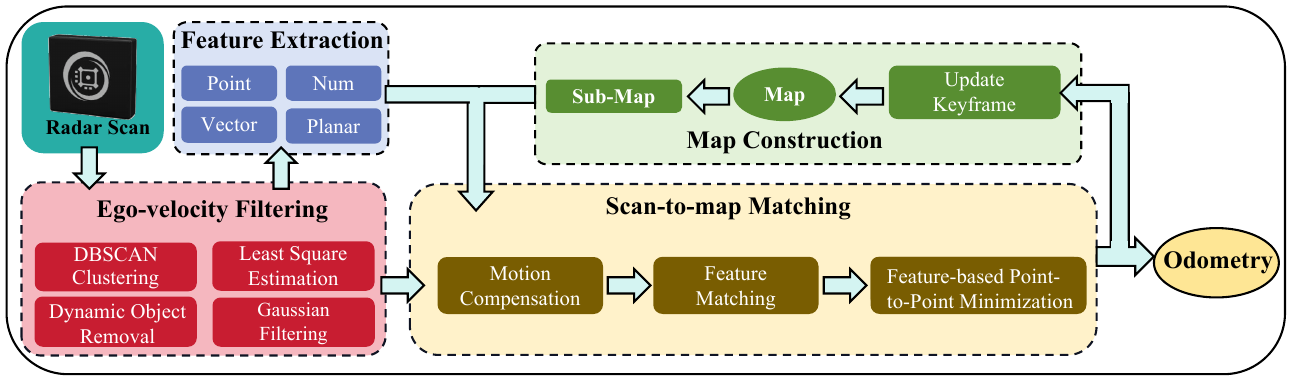}
\caption{Framework Overview of EFEAR-4D.}
\label{framework}
\vspace{-2pt}
\end{figure*}
To tackle these challenges, this paper introduces a novel method for 4D radar odometry estimation, \textbf{EFEAR-4D} (Ego-velocity Filtering for Efficient and Accurate 4D Radar-odometry), as shown in Fig. \ref{fig1}. Specifically, we associate direction with Doppler velocity for each point and use a density-based clustering method DBSCAN \cite{schubert2017dbscan} to filter out noises and remove point cloud of dynamic objects, followed by ego-velocity estimation leveraging the geometric distribution characteristics of Doppler velocity from all static objects. Considering sparsity in radar point cloud, we conduct region-wise geometric feature extraction for registration. And Point-to-Point(P2P) registration is employed for computing odometry. Moreover, To investigate how limited FOV of 4D radars affects the numbers of point cloud and thus the performance of odometry estimation, we design experiments with 4D radar positioned at various heights, and provide comparative analysis as a guidance for future 4D radar real-world deployments. 
Our main contributions are summarized as follows:  
\begin{enumerate}
\item A novel 4D radar odometry framework, a combination of Doppler velocity, feature-based P2P matching, is introduced. This framework effectively addresses the challenge of sparsity in 4D radar point cloud, enabling robust and precise localization in different urban environments;
\item An accurate method for estimating ego velocity is presented, exploiting the proposed physical model of radial Doppler velocity. Clustering and signal pre-processing algorithms are utilized to filter out dynamic objects and noises,  maintaining robust performance even in the presence of complex dynamic scenes;
\item The proposed odometry method EFEAR-4D is thoroughly validated across various scales, vehicles, and environments, consistently demonstrating high performance in diverse scenarios. A dataset with 4D radar installation at different heights is complied. The dataset and code are open-sourced to facilitate related research in the community.
\end{enumerate}

\section{Related Works}
We will briefly summarize related works from two aspects: traditional radar (2D, including the $x$ and $y$ axes, or 3D, including the $x, y$ axes, and Doppler velocity) and 4D imaging radar (including the $x, y, z$ axes and Doppler velocity) odometry.
\subsection{Traditional radar odometry} 
Radar-based odometry methods mostly can be divided into two categories: feature-based and direct methods.  Feature-based methods typically extract and match distinctive features for registration between consecutive scans. Cen et al. \cite{8460687} recognize landmarks as features to perform data association.  In radarSLAM \cite{hong2022radarslam}, SURF descriptor is extracted for pose estimation.  Adolfsson et al. \cite{adolfsson2022LiDAR} incorporate normal vector  and other factors as weights for point-to-line scan matching. For direct methods, \cite{checchin2010radar} uses the Fourier-Mellin transform to recover all rigid transformation parameters. ICP and NDT are used for radar scan registration in \cite{holder2019real} and \cite{rapp2017probabilistic}, respectively.  
\subsection{4D radar odometry}
The Doppler velocity information from 3D/4D radars offers a novel method to estimate ego velocity, which traditional 2D radars cannot achieve. Recently, several works have proposed Doppler velocity estimation algorithms for spinning radars \cite{lisus2024doppler} \cite{rennie2023doppler}. However, the extension of these methods to ego velocity estimation remains an open research question.
 With accurate ego-velocity estimation as a reliable motion prior, pose estimation can be enhanced significantly.
The simple idea of ego-velocity estimation is to consider the Doppler velocity of each stationary point as a projection of the true ego-velocity in the radial direction between the stationary point
and the center of the radar. Consequently, the least squares methods can be used to calculate the ego-velocity. However, substantial noises emerge in 4D radar velocity measurements, particularly in scenarios involving dynamic objects, where the obtained Doppler velocity represents the relative velocity between two moving entities. 

Kramer et al. \cite{kramer2020radar} introduce Cauchy robust norm to address this challenge. Christopher and Gert \cite{doer2020ekf} utilize RANSAC method to mitigate the effects of outliers. 
Though it works well with both small and large numbers of points \cite{kubelka2024we}, its performance deteriorates in complex environments with crowded dynamic objects, due to the lack of identifying and removal of dynamic objects in the point cloud.
Chen et al. \cite{chen2023drio} present a 2D ego-velocity estimation method by refining ground points, however, such a method does not work in 4D cases. Due to sparsity in 4D radar point cloud, many frames lack sufficient ground points. Zeng et.al. \cite{zeng2023joint} present a signal model for radar ego-motion estimation in the presence of velocity ambiguity. However, it concentrates on the fuzziness of the radar point cloud itself rather than the impact of factors such as dynamic objects in real-world situations. Therefore, we aim 
 at proposing a novel ego-velocity filtering method, which provides accurate ego-velocity estimates by filtering out significant noises in diverse dynamic scenarios.

After an accurate ego-velocity estimation, it comes to another process of registration. For 4D radar scan registration, a natural idea is to re-use methods in traditional radar or LiDAR SLAM with appropriate adjustments. APDGICP \cite{zhang20234DRadarSLAM}, a variant from generalized ICP (GICP) \cite{segal_generalized-icp_2010},  exhibits promising performance in certain low-speed campus environments. Nevertheless, its effectiveness deteriorates significantly in high-speed driving scenarios. Li et al. \cite{li20234d} utilize a sliding window to build a denser radar submap and then use NDT \cite{magnusson2009three} for matching. 
In this letter, we present a feature-based registration method adapted from CFEAR \cite{adolfsson2022LiDAR} 2D radar odometry and apply it to 4D radar, achieving high performance in complex environments.
\section{Methodology}
An overview of 4D radar Odometry is shown in Fig. \ref{framework}. 
In this section, we detail the components including ego-velocity filtering for prior guess (\ref{pre-process}), region-wise geometric feature extraction (\ref{feature-extraction}), and a scan-to-map matching based on motion compensation (\ref{s2m}).
\subsection{Ego-velocity filtering}\label{pre-process}
In this section, we aim to compute an accurate ego velocity as a prior guess for subsequent components of odometry estimation. The pipeline of ego-velocity filtering consists of four steps: first  Gaussian filtering to filter out noises in the raw point cloud, followed by DBSCAN clustering static and dynamic points separately, then removing dynamic objects, and finally estimating ego velocity using static points. 

Doppler velocity is the relative velocity in the radial direction from the center of the radar to the measured point. Given that the measured point is a stationary point, we have:
\begin{equation} \label{dv}
|\textbf{v}^{R}|=\textbf{v}^{S}\cdot \textbf{r} = |\textbf{v}^{S}|\cos\theta
\end{equation}
where  $\textbf{v}^R \in \mathbb{R}^3$ is the Dopper velocity (in the radial direction from the center of the radar to the measured point) and  $\textbf{v}^S \in \mathbb{R}^3$ is the ego-velocity vector, $\textbf{r} \in \mathbb{R}^3$  is a unit vector along the direction of the observed point in the radar coordinate system, indicating the direction of the Doppler velocity, and $\theta$ is the angle from $\textbf{v}^R$ to $\textbf{v}^S$, ranging from $-\pi$ to $\pi$.

\begin{figure}
    \centering
\includegraphics[width=0.45\textwidth]{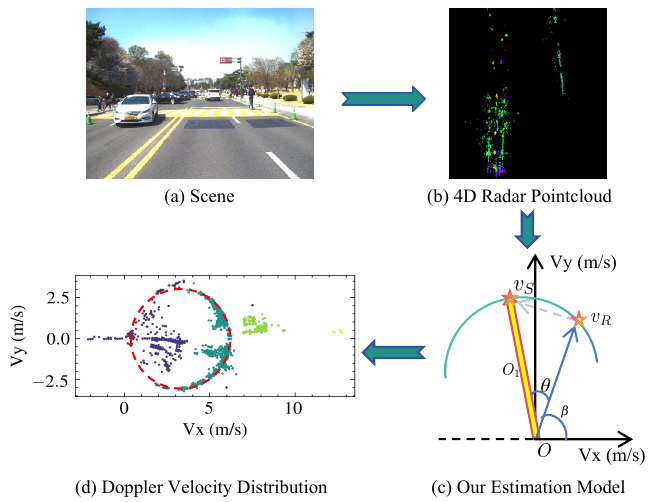}
\caption{The implementation of ego-velocity estimation. In (d), \textcolor[HTML]{006633}{dark green points} are stationary points. \textcolor[HTML]{33CC00}{light green points} represent the white car toward to the camera. And others are other dynamic objects or noise.}
\label{velocity}
\end{figure}

Given that the velocity along the 
z-axis is negligible in ground vehicle settings, we simplify the model by directly setting the z-axis velocity to zero. As depicted in Fig. \ref{velocity} (c),  in the default radar body frame $V_x$-$V_y$-$O$, denote the angle from the x-axis to $\textbf{v}^R$ as $\beta$. and  the decomposing component of $\textbf{v}^R$ in the $V_x$ and $V_y$ coordinates as $v_x, v_y$, respectively. It follows from \eqref{dv} that:
\begin{align*}
    v_x &= |\textbf{v}^R|\cos \beta =\frac{|\textbf{v}^S|}{2}(\cos{(\theta +\beta)}+\cos{(\beta-\theta)})\\
    v_y &= |\textbf{v}^R|\sin \beta   =\frac{|\textbf{v}^S|}{2}(\sin{(\theta +\beta)}+\sin{(\beta-\theta)})
\end{align*}
Defining $\gamma \triangleq \beta + \theta$, then it follows that \footnote{Noting that $\textbf{v}^S$ is on the circle, the other candidate equation $
   \Big(v_x-\frac{|\textbf{v}^S|}{2}\cos{(\beta-\theta)}\Big)^2
   + \Big(v_y-\frac{|\textbf{v}^S|}{2}\sin{(\beta-\theta)}\Big)^2= \Big(\frac{|\textbf{v}^S|}{2}\Big)^2$ is excluded.}: 
\begin{equation}\label{equ:circle}
   \Big(v_x-\frac{|\textbf{v}^S|}{2}\cos{\gamma}\Big)^2
   + \Big(v_y-\frac{|\textbf{v}^S|}{2}\sin{\gamma}\Big)^2= \Big(\frac{|\textbf{v}^S|}{2}\Big)^2
\end{equation}

From a geometric perspective, the above equation \eqref{equ:circle} indicates that all 2D points $(v_x,v_y)$ lie on the same circle with $(\frac{|\textbf{v}^S|}{2}\cos{\gamma}, \frac{|\textbf{v}^S|}{2}\sin{\gamma})$ as the center of circle and  $|\textbf{v}^S|$ as the diameter,  as illustrated in Fig. \ref{velocity} (d).  
Thus $\textbf{v}^S=\begin{bmatrix}
   |\textbf{v}^S|\cos{\gamma},|\textbf{v}^S|\sin{\gamma},0 
\end{bmatrix}^\top$.  Referring to  \eqref{equ:circle},  we can employ the least squares method to compute $\textbf{v}^S$
with the minimization function:
\begin{equation} \label{ls}
    \mathop{\arg\min}\limits_{x_0, y_0} \sum_{i} [ (v^{i}_x-x_0)^2+(v^{i}_y-y_0)^2-(x_0^2+y_0^2)]^2
\end{equation}
where $v^{i}_x$ and $v^{i}_y$ are the $x, y$ components of Doppler velocity from the $i$-th point in the radar point cloud.

After obtaining  optimal solution $x^*_0$ and $y^*_0$ from \eqref{ls}, it follows that: $$\textbf{v}^S=\begin{bmatrix}   |\textbf{v}^S|\cos{\gamma},|\textbf{v}^S|\sin{\gamma},0 
\end{bmatrix}^\top=\begin{bmatrix}
  2x^*_0, 2y^*_0, 0  
\end{bmatrix}^\top.$$ 
However, due to the presence of dynamic objects and noises in radar point cloud, applying the least squares method directly to all data cannot yield correct ego velocity estimates. As illustrated in Fig. \ref{velocity}(d), the \textcolor[HTML]{33CC00}{light green points} correspond to the white car driving in the opposite direction of the ego vehicle, and the dark points correspond to the pedestrians, the black car in front and some other moving objects shown in Fig. \ref{velocity}(a). To differentiate between dynamic points, noise, and stationary points, we employ DBSCAN \cite{schubert2017dbscan}, a density-based clustering method that utilizes region growing on all pairs of $(v_x, v_y)$, the distribution of Doppler velocity in the horizonal plane. Although noise and dynamic objects may not be entirely distinguishable, stationary points will dominate, allowing for the straightforward identification of stationary points and facilitating more accurate estimation.  Dynamic points are then removed according to the classification results of DBSCAN for better registration in the subsequent steps.

\subsection{Region-wise feature extraction}\label{feature-extraction}
Since the point cloud of 4D radar is noisy, direct registration with all the points would lead to large errors, especially in the $z$-axis. As such, we adapt the novel 2D radar odometry method CFEAR \cite{adolfsson2022LiDAR} to the characteristics of 4D radar.  Following the procedure of CFEAR,  the initial step is to partition the filtered points into voxels. For each voxel $V_i$, then a KD-tree \cite{ram2019revisiting} is used to search for neighbouring points, including points both within and adjacent to the voxel, denoted as $P_i$. Subsequently, the weighted mean point for $P_i$ is calculated using intensity weights, which serves as the feature point of the voxel. The weighted sample covariance matrix is denoted as $C_i$ and the feature point for $P_i$ is denoted as $p_i$. 

\begin{figure}
    \centering
    \includegraphics[width=\linewidth]{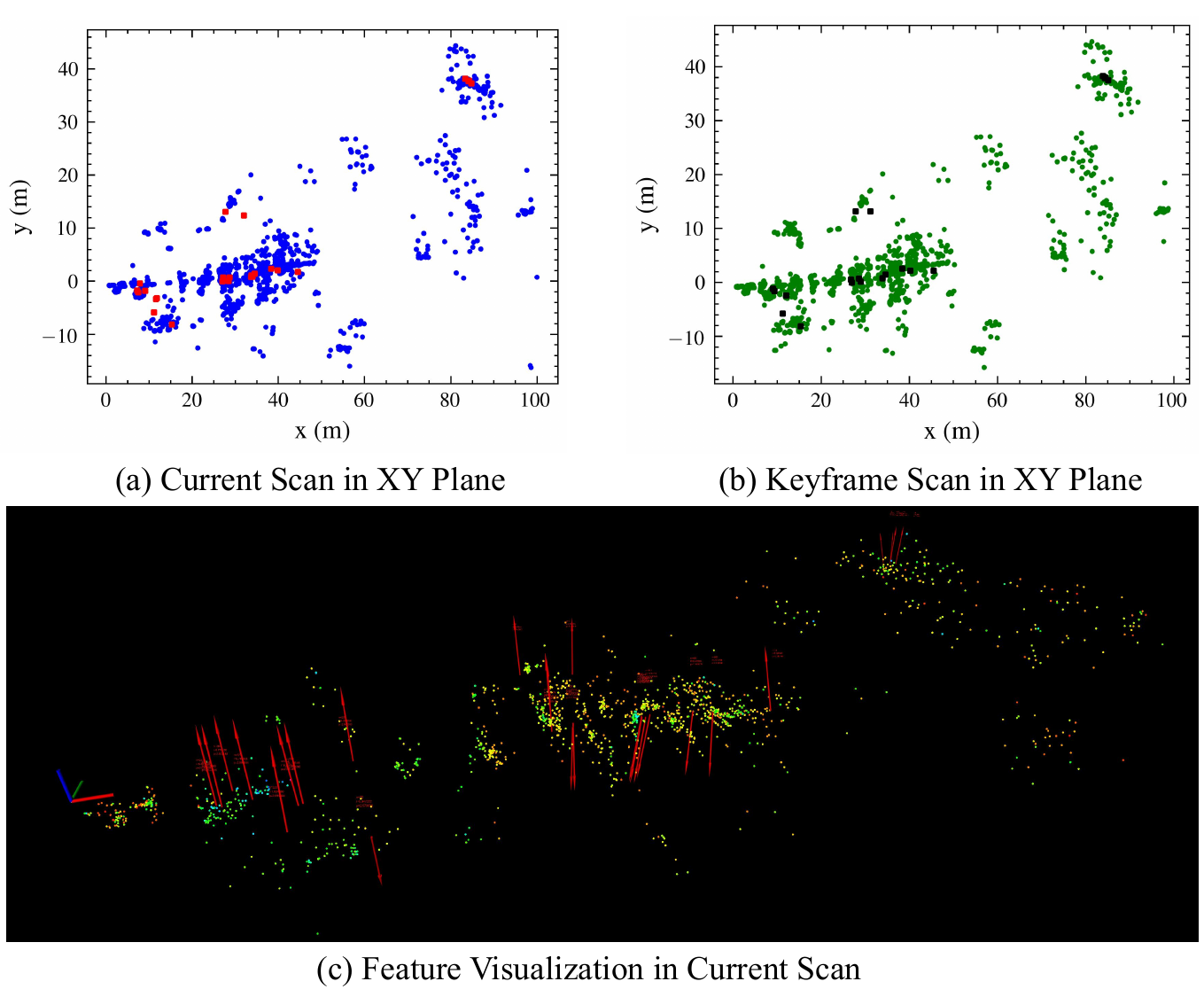}
    \vspace{-1.5em}
    \caption{The current scan (a) and corresponding keyframe (b) are projected in the XY plane. \textcolor[HTML]{FF0000}{Red points} in (a) and black points in (b) are successfully matched feature points in the two scans, respectively. The real current point cloud and extracted normal vectors are shown in (c). For visualization, the size of voxels here is 10m, with features being much denser in quantity during actual operation.}
    \label{fig:feature}
\end{figure}

Moreover, to improve noise immunity, we perform principal component analysis (PCA) \cite{mackiewicz1993principal} to compute normal vector $u_i$ as another feature of the voxel $V_i$. The eigenvalues and eigenvectors of  covariance matrix $C$ are calculated as follows:
\begin{equation}
    C_{\alpha} u_{\alpha} = \lambda_{\alpha}  u_{\alpha}
\end{equation}
where $\alpha=1,2,3$ and $\lambda_1 > \lambda_2 > \lambda_3$. The smallest eigenvector represents $u_i$. The normal vectors in one scan are visualized in Fig. \ref{fig:feature} (c). The condition number $\kappa =\lambda_1 /\lambda_3$  is also calculated as a feature. To some extent, a higher value of $\kappa$ indicates greater stability and reliability of the features within this voxel. 


\begin{table*}[htbp]
    \centering
    \caption{  QUANTITATIVE ANALYSIS: RMSE of Absolute Trajectory Error (ATE) on MSC Dataset}
   \resizebox{0.99\textwidth}{!}{
    \begin{tabular}
    {ccccc ccccc cccccc ccccc}
    \hline
         &\multirow{2}{*}{\diagbox[width=8em]{\textbf{Method}}{\textbf{Sequence}}}
         &\multirow{2}{*}{URBAN\_{}A0}&\multirow{2}{*}{URBAN\_{}B0}&\multirow{2}{*}{URBAN\_{}C0}&\multirow{2}{*}{URBAN\_{}D0}&\multirow{2}{*}{URBAN\_{}D1
 }&\multirow{2}{*}{URBAN\_{}F0}&\multirow{2}{*}{RURAL\_{}A2}&\multirow{2}{*}{RURAL\_{}D2}\\
 \\
        \hline 
        LiDAR&
        LEGO-LOAM \cite{shan2018lego} &3.1859&33.7488&0.8671&\textbf{4.4209}&\textbf{2.9539}&9.9085&\textbf{7.9991}&1.0396 \\
         Vision& ORB-SLAM3 \cite{campos2021orb}&11.6498&18.0216&2.2948&49.0342&52.0378&31.2399&74.3235&1.6097\\
         \hdashline
         \multirow{4}{*}{4D radar}
         &4DRadarSLAM-apdgicp \cite{zhang20234DRadarSLAM}&10.1287&39.7497&13.8928&104.1306&42.6255&23.8007&65.4221&0.8184\\
         &4DRadarSLAM-gicp \cite{zhang20234DRadarSLAM}&7.7078&7.6002&6.1576&92.0331&30.5685&13.5864&42.3049&\textbf{0.5769}\\
        & Ours-GICP&6.5796&3.4610&4.4425&57.8950&17.1938&10.9256&39.2468&1.0908\\
         &Ours &\textbf{2.4282}&\textbf{3.2039}&\textbf{0.8566}&5.0714&8.2494&\textbf{5.9241}&16.4847&1.1431\\
         \hline
    \end{tabular}
    }
    \label{result_msc}
    \vspace{-1em}
\end{table*}

\subsection{Scan-to-map matching}\label{s2m}
Since the 4D radar point cloud are sparse, it is impossible to estimate poses reliably by matching only two consecutive frames. Therefore, we construct a keyframe sliding window for registration. Keyframes are selected based on pose deviation exceeding a predetermined threshold in rotation or translation from the preceding keyframe. The map points of keyframe within the sliding window collectively form a sub-map utilized directly in subsequent steps. 

Utilizing the ego-velocity calculated in the previous step, we derive the initial pose estimate $ ^{w}\hat{T_k}$ based on a constant velocity model and align it with the sub-map coordinates for the current frame, denoted as $^{s}\hat{T_{k}}$. Therefore, if the velocity estimation is accurate, the coordinates of the same reflective points in the sub-map and the current frame should be close. Then each voxel in the current frame can search for its corresponding voxel in the sub-map in a certain radius. For corresponding voxels, feature points are then matched and a filtering process is applied to exclude points with a normal vector dot product $u_i \cdot u_j < z_{thre}$, indicating potential false association. As a result, all matching points ($\prescript{s}{}{p}, \prescript{k}{}{p}$) form a sparse set $\Omega$ as shown in Fig. \ref{fig:feature} (a) and (b).

 For every tuple in $\Omega$, we conduct a P2P matching. We set different weights for every tuple to distinguish some stable matching and other unstable points. The weights come from the similarity  of features selected in \ref{feature-extraction} and can be calculated as follows: 
 \begin{equation}
     w_i = w^{vec}_i+w^{con}_i+w^{num}_i 
 \end{equation}
where $w^{vec}_i$,$w^{con}_i$ and $w^{num}_i$ represents the similarity of normal vectors, condition numbers, and number of points in $\prescript{s}{}{P_i}$ and  $\prescript{k}{}{P_i}$, respectively.  Similar to \cite{adolfsson2022LiDAR}, here we use Huber loss as loss function due to its robustness in handling outliers. The minimization function can be formulated as follows:
\begin{equation}
    \mathop{\arg\min}\limits_{^sT_k}\sum_{i \in C} w_i L_{\delta}(||\textbf{e}_i||^2)
\end{equation}
where $\textbf{e}_i= \prescript{s}{}{p}_i-(^sT_k \prescript{k}{}{p}_i)$.
\section{Experiments}\label{section:Exp}


\subsection{Quantitative Evaluation}

\subsubsection{\textit{Comparative evaluation with MSC dataset}}
\begin{figure}[t]
\begin{minipage}[t]{0.115\textwidth}
\centering
\includegraphics[width=1\textwidth]{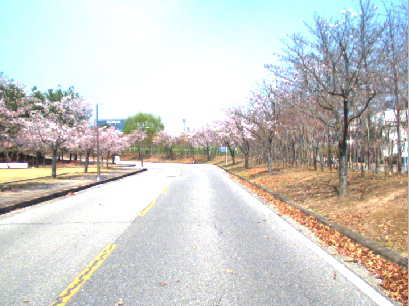}
\centerline{(a) U\_{}A$0^{*}$}
\end{minipage}
\begin{minipage}[t]{0.115\textwidth}
\centering
\includegraphics[width=1\textwidth]{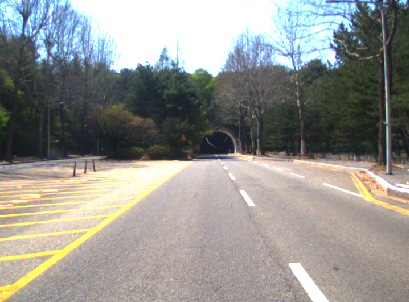 }
\centerline{(b) U\_{}B0}
\end{minipage}
\begin{minipage}[t]{0.115\textwidth}
\centering
\includegraphics[width=1\textwidth]{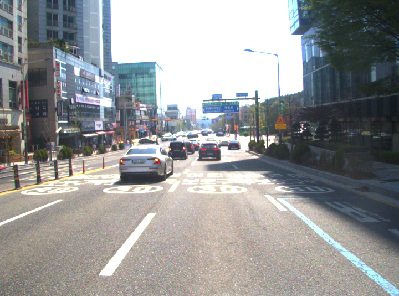 }
\centerline{(c) U\_{}C0}
\end{minipage}
\begin{minipage}[t]{0.115\textwidth}
\centering
\includegraphics[width=1\textwidth]{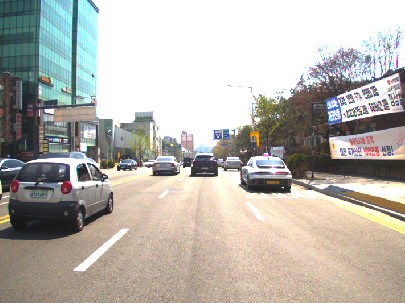 }
\centerline{(d) U\_{}D0}
\end{minipage}

\vspace{0.4em}

\begin{minipage}[t]{0.115\textwidth}
\centering
\includegraphics[width=1\textwidth]{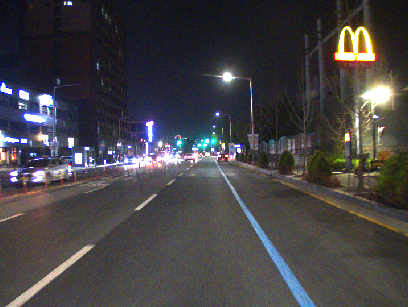 }
\centerline{(e) U\_{}D1}
\end{minipage}
\begin{minipage}[t]{0.115\textwidth}
\centering
\includegraphics[width=1\textwidth]{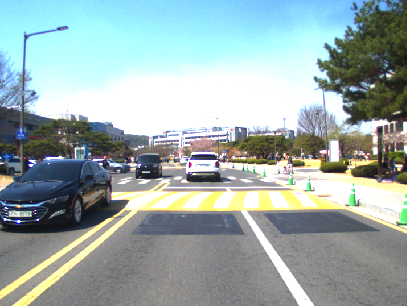 }
\centerline{(f) U\_{}F0}
\end{minipage}
\begin{minipage}[t]{0.115\textwidth}
\centering
\includegraphics[width=1\textwidth]{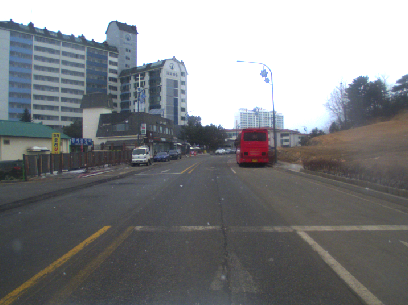 }
\centerline{(g) R\_{}A$2^*$}
\end{minipage}
\begin{minipage}[t]{0.115\textwidth}
\centering
\includegraphics[width=1\textwidth]{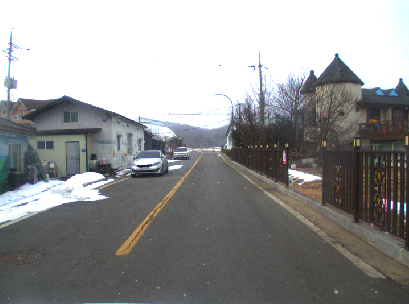 }
\centerline{(h) R\_{}D2}
\end{minipage}
\caption{Scenes of different MSC dataset sequences. "*" means the sequence is a loop.}
\label{msc_scene}
\vspace{-1em}
\end{figure}

We start by comparing our method to previously published odometry and SLAM methods on the MSC dataset \cite {choi2023msc}. Our comparative study encompasses several vision, LiDAR, and 4D radar SLAM baselines. Specifically, ORB-SLAM3 \cite{campos2021orb}, a widely recognized visual SLAM system, and LEGO-LOAM \cite{shan2018lego}, a LiDAR-based SLAM approach, are selected for comparison.  For fair comparison, we use only a stereo camera for ORB-SLAM3 as all baselines employ a single sensor, focusing solely on comparing odometry. For radar, we compare with 4DRadarSLAM \cite{zhang20234DRadarSLAM}, the only open-source algorithm in 4D radar slam, including two options: GICP and APDGICP. Our evaluation primarily revolves around Absolutely Trajectory Error (ATE) and Relative Pose Error (RPE) metrics \cite{grupp2017evo}, with RTK serving as the ground truth. The corresponding scenes are depicted in Fig. \ref{msc_scene}, encompassing static (U\_{}A0, U\_B0{}), high-dynamic (U\_{}C0, U\_{}D0, U\_{}F0), high-speed (U\_{}D0, U\_{}D1), night (U\_{}D1), snowy (R\_{}A2, R\_{}D2), and loop environments (U\_{}A0, R\_{}A2). 

\begin{figure*}[t]
\begin{minipage}[t]{0.33\textwidth}
\centering
\includegraphics[width=1\textwidth]{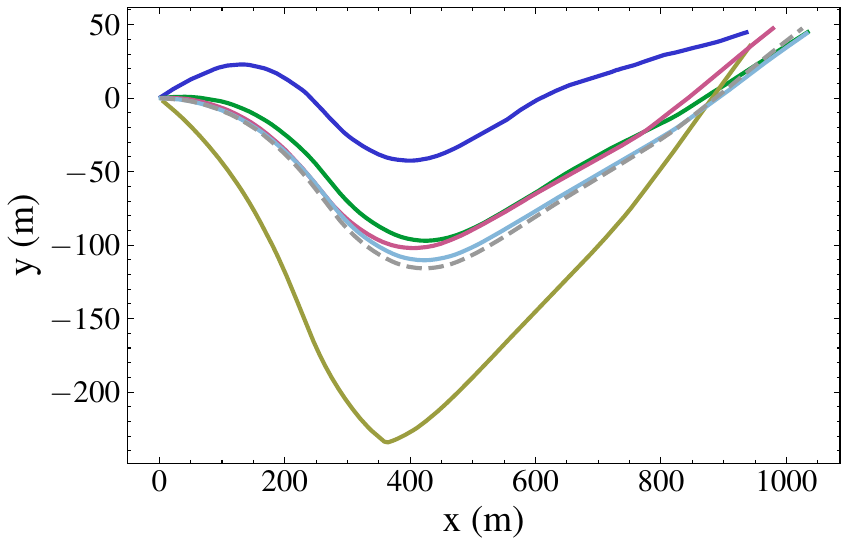 }
\end{minipage}
\begin{minipage}[t]{0.33\textwidth}
\centering
\includegraphics[width=1\textwidth]{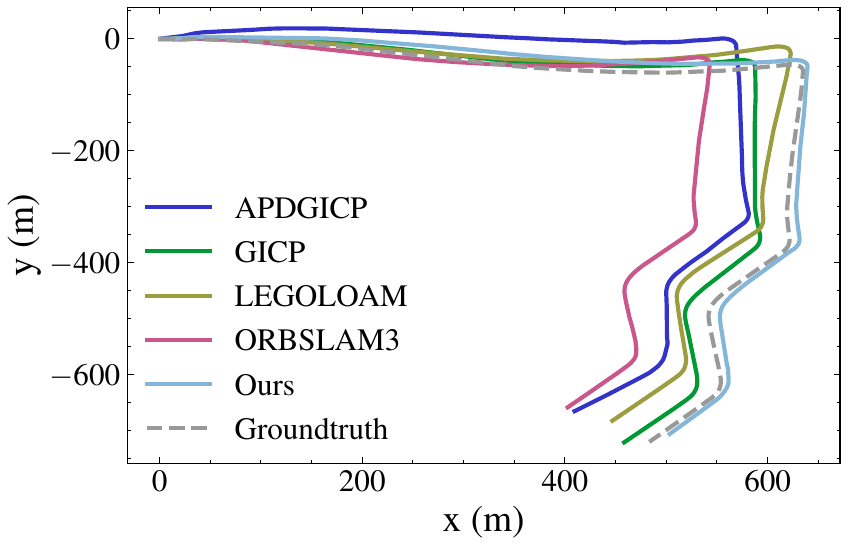 }
\end{minipage}
\begin{minipage}[t]{0.33\textwidth}
\centering
\includegraphics[width=1\textwidth]{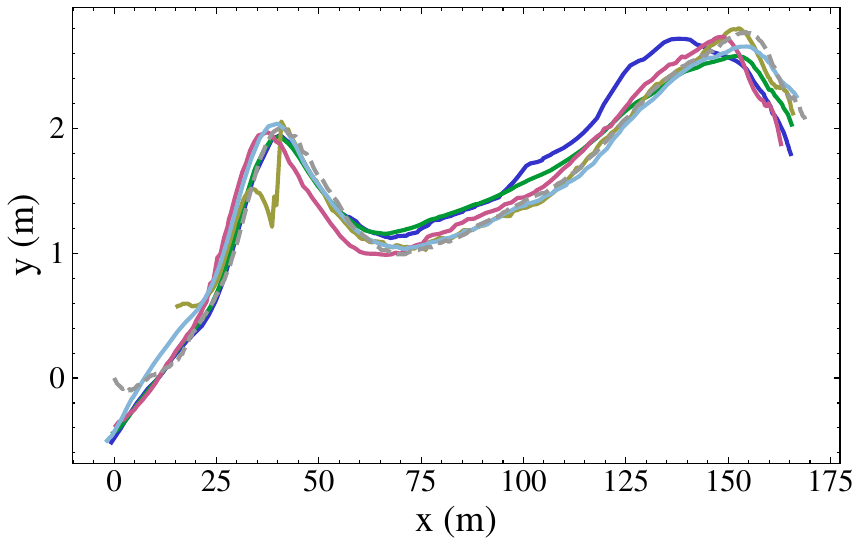 }
\end{minipage}

\vspace{0.2em}

\begin{minipage}[t]{0.33\textwidth}
\centering
\includegraphics[width=1\textwidth]{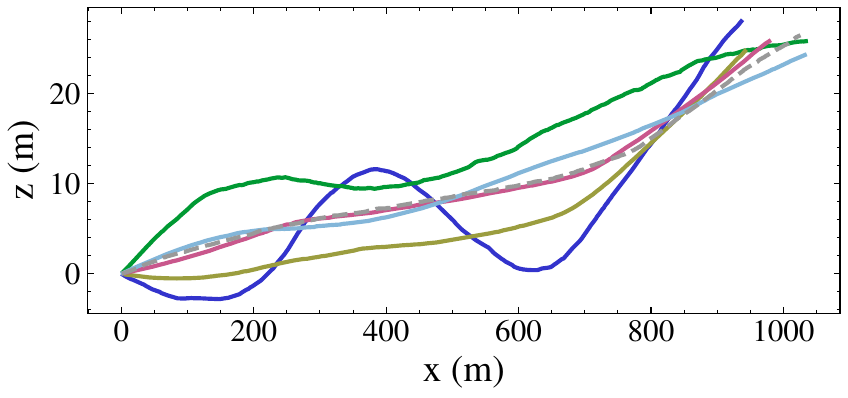 }
\centerline{(a) URBAN\_{}B0}
\end{minipage}
\begin{minipage}[t]{0.33\textwidth}
\centering
\includegraphics[width=1\textwidth]{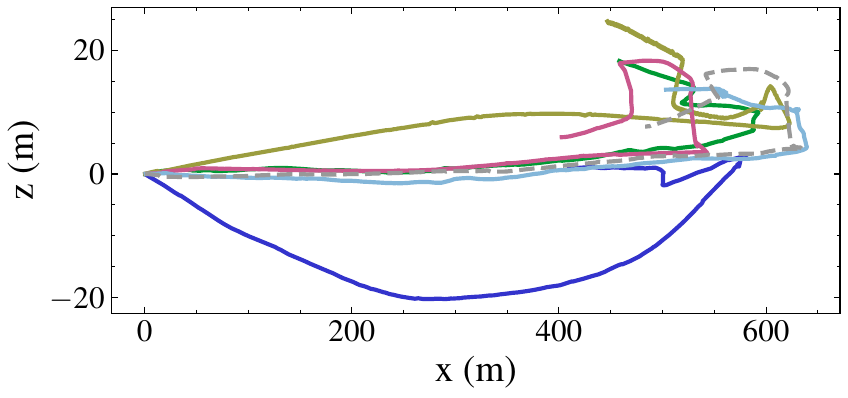 }
\centerline{(b) URBAN\_{}F0}
\end{minipage}
\begin{minipage}[t]{0.33\textwidth}
\centering
\includegraphics[width=1\textwidth]{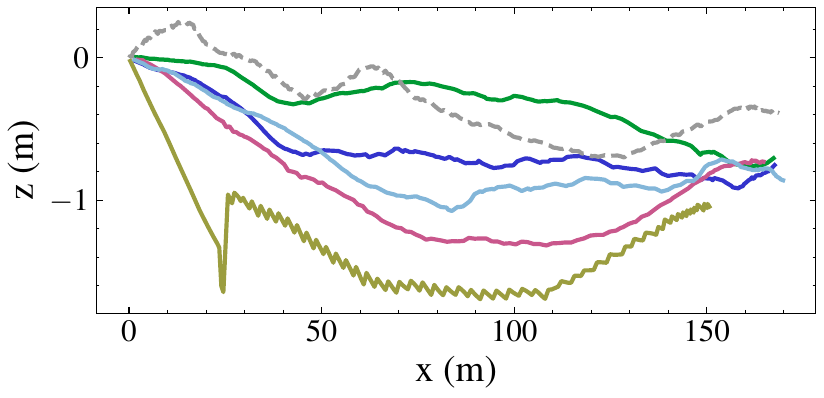 }
\centerline{(c) RURAL\_{}D2}
\end{minipage}
\caption{Trajectories for six sequences using different SLAM algorithms on MSC dataset.}
\label{msc_rpg}
\end{figure*}

\begin{figure}
\vspace{-1.5em}
    \centering
 \begin{minipage}[t]{0.49\linewidth}
\centering
\includegraphics[width=1\linewidth]{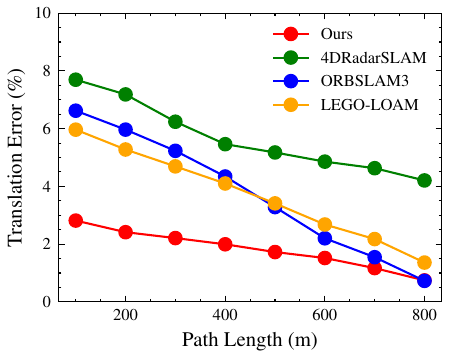}
\end{minipage}
 \begin{minipage}[t]{0.49\linewidth}
\centering
\includegraphics[width=1\textwidth]{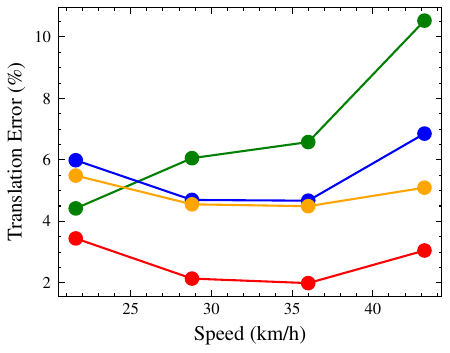}
\end{minipage}

\centerline{(a) URBAN\_{}A0}

 \begin{minipage}[t]{0.49\linewidth}
\centering
\includegraphics[width=1\linewidth]{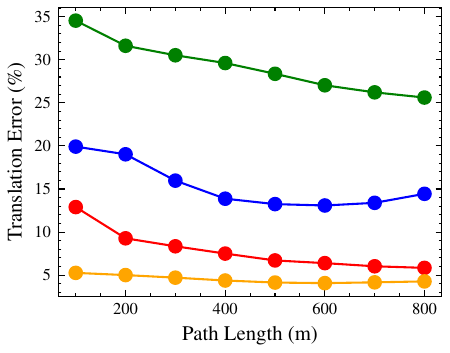}
\end{minipage}
 \begin{minipage}[t]{0.49\linewidth}
\centering
\includegraphics[width=1\textwidth]{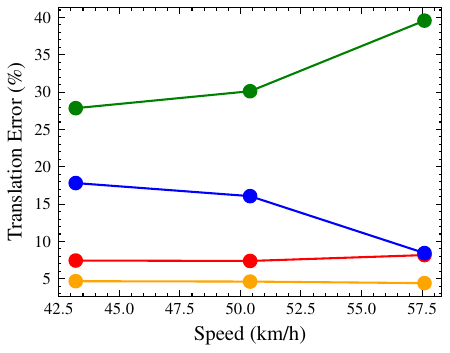}
\end{minipage}

\centerline{(b) URBAN\_{}D1}

 \begin{minipage}[t]{0.49\linewidth}
\centering
\includegraphics[width=1\linewidth]{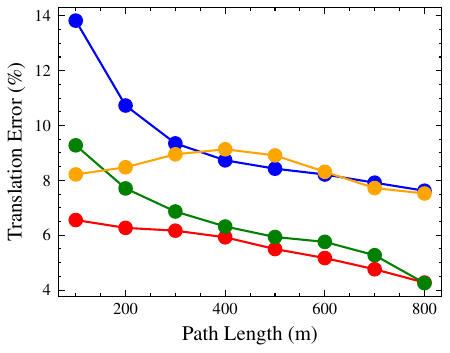}
\end{minipage}
 \begin{minipage}[t]{0.49\linewidth}
\centering
\includegraphics[width=1\textwidth]{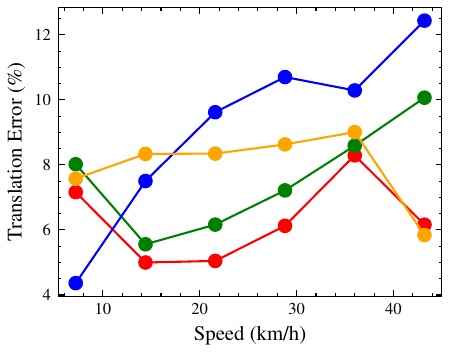}
\end{minipage}

\centerline{(c) URBAN\_{}F0}
\caption{Relative error of different path lengths and speeds on some sequences of MSC dataset.}
\label{ts_tl}
\vspace{-1em}
\end{figure}
We present the odometry error in Tab. \ref{result_msc} and depict the trajectory results  in Fig. \ref{msc_rpg}. Unexpectedly, we observe that our errors are lower than those of ORB-SLAM3 across nearly all scenarios and are approaching the level of LiDAR, or even surpassing it in some scenarios.
In static environments (U\_{}A0, U\_{}B0), our EFEAR-4D method achieves accurate pose estimation due to precise ego-velocity estimation, outperforming other methods. In sequence U\_{}B0, dense vegetation challenges LiDAR, degrading LEGO-LOAM's performance. Other radar methods show significant errors due to lack of feature extraction. In high-speed environments, rapid motion causes large gaps between the consecutive point cloud, reducing features for matching. Our method maintains low error rates due to accurate ego-velocity estimation. LEGO-LOAM's larger FOV results in better stability and performance than radar and camera methods, such as ORB-SLAM3, which suffer from limited FOV. In dynamic environments, our method minimizes the impact of dynamic objects by removing them and leveraging weak reflections, outperforming vision-based and LiDAR-based methods. It also outperforms GICP and APDGICP by extracting features and filtering unstable points. In snowy environments, particularly in R\_{}D2, radar-based methods outperform others as LEGO-LOAM occasionally fails to register frames, demonstrating radar's efficacy in all weather conditions.

Fig. \ref{ts_tl} describes the relative error of ORB-SLAM3, LEGO-LOAM, baseline radar odometry, and our radar odometry EFEAR-4D using different path lengths and speeds with different sequences, following the popular KITTI odometry evaluation protocol \cite{geiger2012we}. Our EFEAR-4D shows a stable performance on all the datasets except in high-speed scenarios where it is slightly inferior to LEGO-LOAM. Both 4DRadarSLAM (apdgicp) and ORB-SLAM3  have a relatively large error and their performances are sensitive to speed variations. It is worth pointing out that both our EFEAR-4D and 4DRadarSLAM perform well in dynamic environments, proving that 4D radar odometry is not less sensitive to moving objects than its counterparts of vision and LiDAR.

\subsubsection{\textit{Comparative evaluation on NTU Dataset}}
\begin{table}[t]
    \centering
    \caption{  QUANTITATIVE ANALYSIS: RMSE of Relative Translation Error and Relative Rotation Error on NTU Dataset}
    \resizebox{0.49\textwidth}{!}{
\begin{tabular}{ccccc ccccc cccccc ccccc}
    \hline
\multirow{2}{*}{\diagbox[width=8em]{\textbf{Sequence}}{\textbf{Method}}}
      &\multicolumn{2}{c}{livox\_{}horizon\_{}loam}
      &\multicolumn{2}{c}{4DRadarSLAM \cite{zhang20234DRadarSLAM}}
      &\multicolumn{2}{c}{Ours}\\
&$t_{rel}$&$r_{rel}$&$t_{rel}$&$r_{rel}$&$t_{rel}$&$r_{rel}$\\
      \hline
      cp&0.0630&0.0195&0.0501&0.0126&0.0509&0.0125\\
      garden&0.0563&0.0107&0.0475&0.0111&0.0379&0.0435\\
      loop1&1.1518&0.0508&0.4300&0.0156&0.3727&0.0152\\
      loop2&0.2217&0.0127&0.3850&0.0174&0.3701&0.0175\\
      nyl&0.0837&0.0129&0.0909&0.0162&0.0893&0.0166\\
      \hline
\end{tabular}
 }
\label{result_ntu}
\end{table}
\begin{figure}
\vspace{-1.5em}
    \centering
    \includegraphics[width=\linewidth]{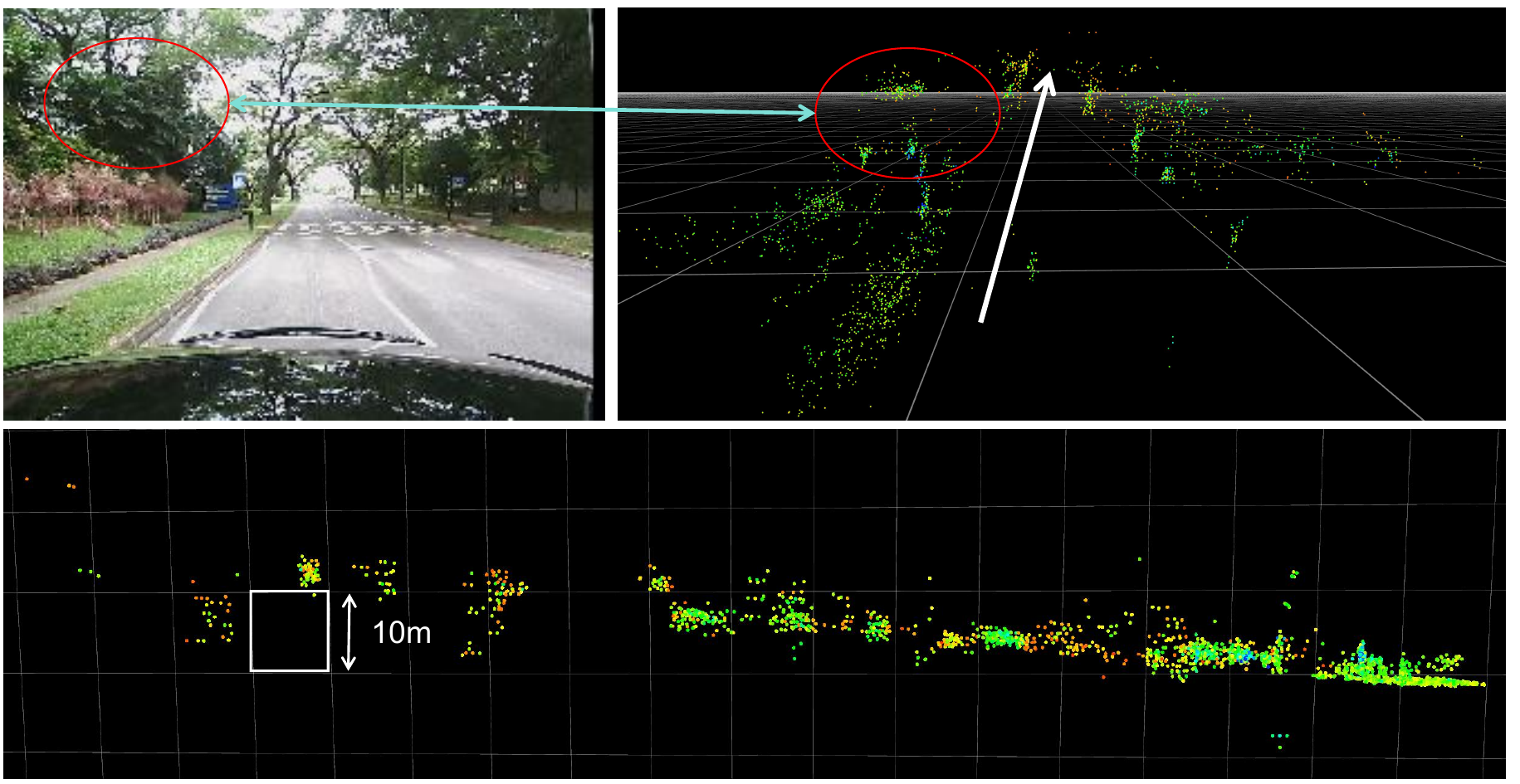}
    \vspace{-1em}
    \caption{The point cloud and corresponding scene in NTU dataset. The majority of points are reflected from trees and bushes along the road.}
    \label{ntu_wrong}
\end{figure}

We compare our methods with 4DRadarSLAM and livox\_{}horizon\_{}loam on NTU dataset \cite{zhang2023ntu4dradlm}. Compared with MSC dataset, the number of sequences on NTU dataset is fewer and the scenes are less challenging since there are no dynamic objects also with relatively low speed. As the dataset only provides Livox horizon LiDAR instead of high-end scanning LiDAR, we use livox\_{}horizon\_{}loam as LiDAR odometry baseline for reference. We evaluate them with the rotation part and the translation part of RPE, respectively. The results are shown in Tab. \ref{result_ntu}. 
In the cp sequence, a low-speed sequence involving travel around buildings, our EFEAR-4D performs well due to its ability to extract stable features such as walls and the ground. However, in the Garden and Nyl sequences, which consist of grassland or bushes, EFEAR-4D performs slightly worse as the extracted features are not sufficiently distinguishable. In Loop 1 and Loop 2, collected by car at moderate speeds, our drifts are also slightly lower than those of 4DRadarSLAM.

Moreover, there exists \textit{a significant difference in the number of points} between the NTU dataset and the MSC dataset, with the features of points in the former appearing more ambiguous and the number of points being much lower. This is because the 4D radar is installed on the roof of the car in the NTU dataset. Owing to the limited FOV of 4D radar, point cloud are formed with reflections of objects in high spaces, such as trees. Most radar lines depicted in Fig. \ref{ntu_wrong} hit trees and reflect back to the radar, exhibiting a higher trend along the road. Therefore, the points exhibit fewer distinguishable features, since trees and bushes are inherently cluttered and lack distinguishable features.

\subsubsection{Ego-velocity estimation evaluation}
\begin{figure}[t]
    \centering
    \includegraphics[width=0.48\textwidth]{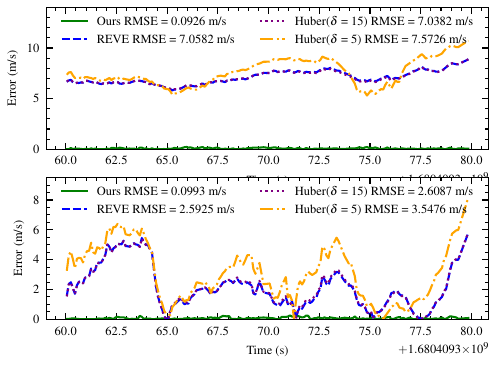}
    \vspace{-0.5em}
    \caption{Ego velocity error comparison between REVE and our method on Dynamic 2 sequence. The figures above and below illustrate the estimation error in the x-direction and y-direction over time, respectively.} 
    \label{line_speed}
    \vspace{-1em}
\end{figure}
\begin{table}[t]
    \centering
    \caption{  QUANTITATIVE ANALYSIS: RMSE of SPEED ERROR ON MSC DATASET}
\begin{tabular}{ccccc ccccc cccccc ccccc}
    \hline
\multirow{2}{*}{\textbf{Environment}}&\multirow{2}{*}{\diagbox[width=8em]{\textbf{Sequence}}{\textbf{Method}}}
      &\multicolumn{2}{c}{REVE \cite{doer2020ekf}}
      &\multicolumn{2}{c}{Ours}\\
      &&$x$&$y$&$x$&$y$\\
      \hline
      \multirow{2}{*}{\textbf{Dynamic}}
      &Dynamic 1&2.2864&2.2921&0.2703&0.8907\\
      &Dynamic 2&7.0582&2.0925&0.0926&0.0993\\
      \multirow{2}{*}{\textbf{Static}}
      &Static 1&0.5264&0.6038&0.2389&0.1027\\
      &Static 2&0.2068&0.3691&0.2055&0.3785\\
      \hline
\end{tabular}
\label{speed_result}
\vspace{-1em}
\end{table}
\begin{figure}[t]
\begin{minipage}[t]{0.24\linewidth}
\centering
\includegraphics[width=1\textwidth]{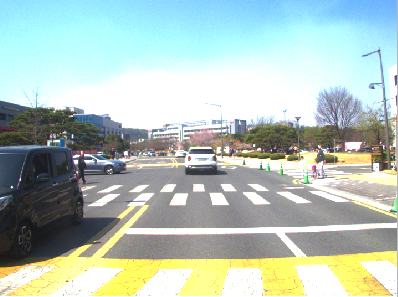 }
\centerline{(a) Dynamic 1}
\end{minipage}
\begin{minipage}[t]{0.24\linewidth}
\centering
\includegraphics[width=1\textwidth]{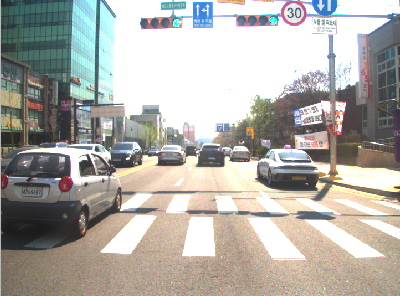 }
\centerline{(b) Dynamic 2}
\end{minipage}
\begin{minipage}[t]{0.24\linewidth}
\centering
\includegraphics[width=1\textwidth]{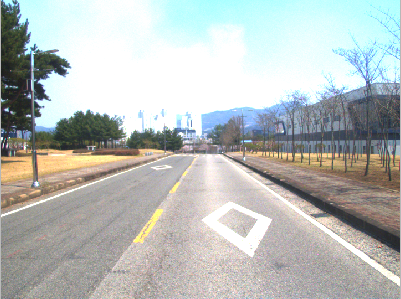 }
\centerline{(c) Static 1}
\end{minipage}
\begin{minipage}[t]{0.24\linewidth}
\centering
\includegraphics[width=1\textwidth]{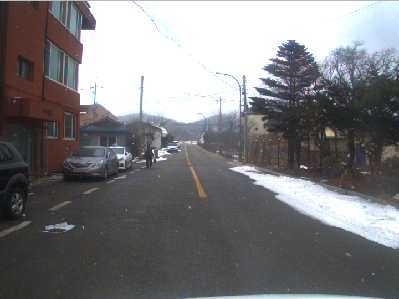 }
\centerline{(d) Static 2}
\end{minipage}
\caption{Scenes of different dynamic and static sequences. Specifically, the Dynamic 1 and Dynamic 2 sequences are extracted from the URBAN\_{}F0 and URBAN\_{}D0 sequences respectively, and the Static 1 and Static 2 sequences are derived from URBAN\_{}A0 and RURAL\_{}D2 in the MSC dataset.}
\label{speed}
\end{figure}
\begin{figure}
    \vspace{-1em}
    \centering
    \includegraphics[width=0.48\textwidth]{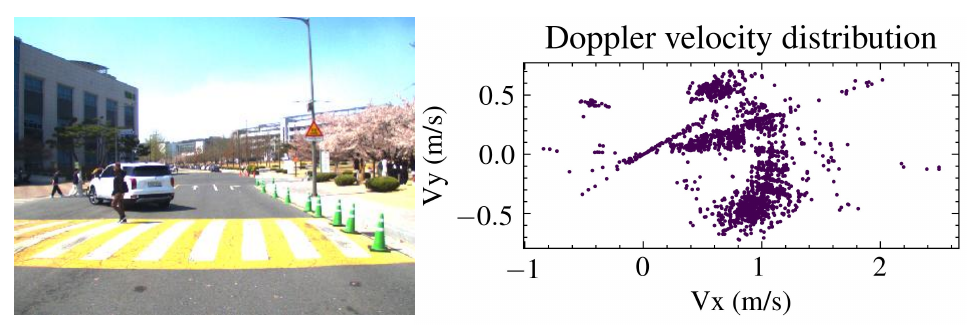}
    \vspace{-0.5em}
    \caption{DBSCAN fails to distinguish dynamic points and static object in some particular complex scenarios. }
    \label{speed_fail}
    \vspace{-1em}
\end{figure}
To further evaluate the performance of our ego velocity estimation, we extract 4 sequences from the MSC dataset, where two are in a dynamic environment, and the other two are in a static environment. Scenes and the specific information about these sequences are illustrated in Fig. \ref{speed}. We compare our method with REVE \cite{doer2020ekf}, an open-source algorithm using RANSAC and linear least square method to compute ego velocity. As shown in Tab. \ref{speed_result}, in dynamic environments, the performance of our method is much better than REVE. The estimation errors of ego velocity in Dynamic 2 sequence are depicted in Fig. \ref{line_speed}. Our ego velocity estimation has not only smaller errors but also smaller jitter. This is because, compared to REVE, we propose and use the physical model in \eqref{equ:circle} that the $x$-axis and $y$-axis components of Doppler velocities from static objects form a circle to estimate ego velocity, which enhances its consistency.

However, we also observe from experiments that DBSCAN may occasionally fail to cluster
in some particularly complex scenarios, as shown in Fig. \ref{speed_fail}. In such a case, all the points are used as one cluster to fit the Doppler distribution circle, resulting in a sub-optimal ego-velocity estimation.
\begin{figure}[t]
    \centering
    \includegraphics[width=\linewidth]{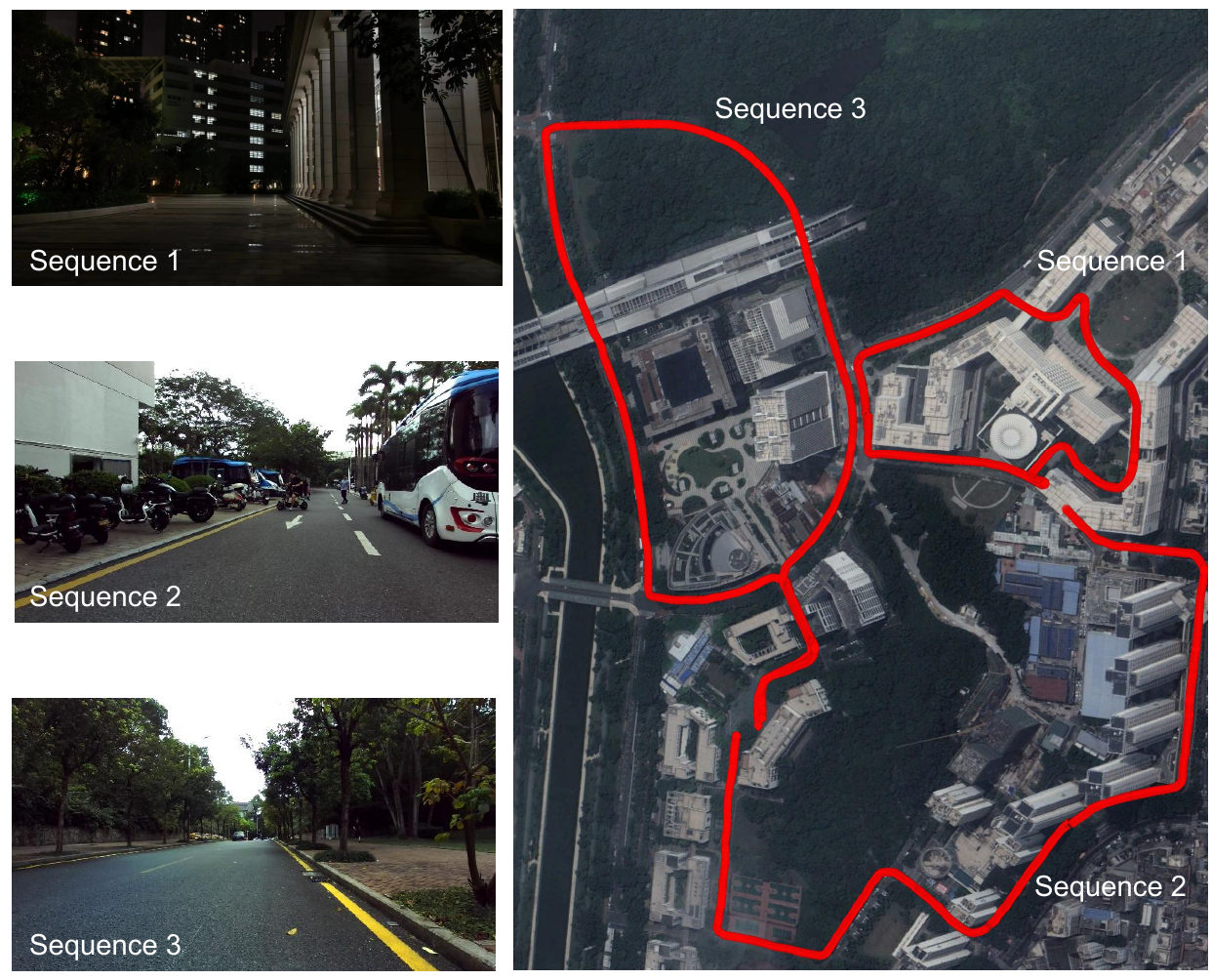}
        \vspace{-1.5em}
    \caption{Scenes and satellite imagery collected in our dataset. The scenes include not only structured environments with dense buildings, but also semi-structured environments with more vegetation and fewer buildings, such as campus main roads. Additionally, the sequences of the same paths with different radar altitudes were collected for further study.}
    \label{scene_realwrold}
    \vspace{-1em}
\end{figure}
\begin{table}[t]
    \centering
    \caption{  QUANTITATIVE ANALYSIS: RMSE of Relative Translation and Rotation Error and Absolutely ERROR on Our Dataset}
    \resizebox{0.49\textwidth}{!}{
\begin{tabular}{ccccc ccccc cccccc ccccc}
    \hline
\multirow{2}{*}{\diagbox[width=8em]{\textbf{Sequence}}{\textbf{Height}}}
      &\multicolumn{2}{c}{High}
      &\multicolumn{2}{c}{Low}\\
&$t_{rel}$&$t_{abs}$&$t_{rel}$&$t_{abs}$\\
      \hline
Sequence1 &0.0529&5.1802&0.0504&11.0891\\
Sequence2 &0.0578&17.9715&0.0536&4.0270\\
Sequence3 &0.0673&3.1901&0.0513 & 3.4317 \\
      \hline
\end{tabular}
 }
 \vspace{-1em}
\label{result_ours}
\end{table}
\begin{figure*}[htbp]
\centering
\begin{minipage}[t]{0.3\linewidth}
    \includegraphics[width=1\textwidth]{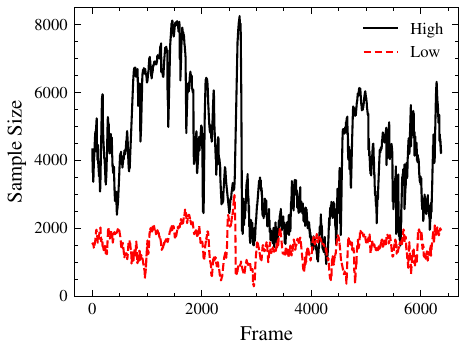}
    \centerline{(a) Sequence 1}
\end{minipage}
\hspace{1em}
\begin{minipage}[t]{0.3\linewidth}
    \includegraphics[width=1\textwidth]{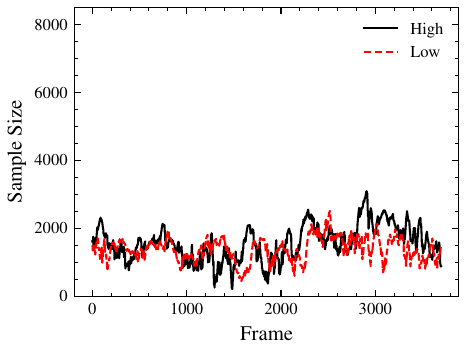}
    \centerline{(b) Sequence 2}
\end{minipage}
\hspace{1em}
\begin{minipage}[t]{0.3\linewidth}
    \includegraphics[width=1\textwidth]{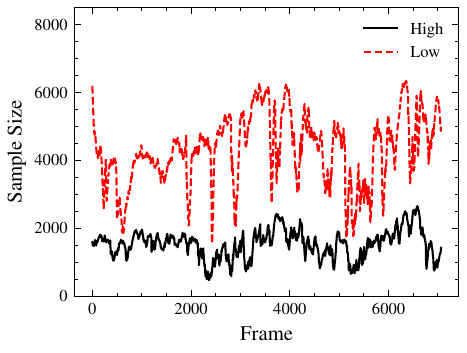}
    \centerline{(c) Sequence 3}
\end{minipage}
\caption{The number of points in different heights with the same route.}
\label{fig_num_our_points}
\vspace{-1.5em}
\end{figure*}

\subsubsection{Self-collected dataset}
\label{dataset}

As mentioned in the previous section, the 4D radar sensor is installed on the roof of the car in the NTU dataset, its height is relatively high (compared to MSC dataset), and the overall number of points in point cloud obtained is lower than that of MSC dataset. To understand the impact of sensor height, we conduct a quantitative study: we place the 4D radar at high and low locations (about 30 cm apart) and collect sequences along the same path by traversing multiple times. Then, we test our algorithm with the same parameters with different radar height settings. 

The route and scenes are depicted in Fig. \ref{scene_realwrold}. All the sequences are collected in University Town, Shenzhen, China. Our mobile robot platform is equipped with an Oculii Eagle 4D radar, configured with a resolution of 0.86m in range, 0.44 deg in azimuth, 0.175 deg in elevation, and 0.27 m/s in Doppler-velocity. Additionally, it features a stereo camera, an IMU, a Robosense LiDAR, and a ByNav GNSS navigation system. Ground truth trajectory is obtained from GPS/IMU + RTK. It is clear to see that the sensor height has a great impact on the number of radar points, as shown in Fig. \ref{fig_num_our_points}. At the same GPS location, putting the radar sensor at different heights results in various signal behaviours in the urban environment. Surprisingly, the number of radar points is inconsistent at different places,  going up and down from place to place.

Sequence 1 captures the vicinity of a teaching building at night. Surrounded by taller buildings, its characteristics remain stable. When the radar is positioned higher, it captures more building-related information, resulting in a denser point cloud. Conversely, radar at lower heights primarily captures adjacent grass areas. Data collected at higher altitudes contain fewer points and discernible features, leading to higher drift along the $z$-axis and increased errors as shown in Table \ref{result_ours}.

In Sequence 2, the point cloud contains few points regardless of radar height, indicating a degraded radar scene with few discernible features. Error rates are lower when the radar is positioned lower, as being closer to the ground allows for more effective data extraction, particularly on highways. In open environments at higher heights, the point cloud reflects off trees, resulting in lack of features. Frames with fewer than 500 points lead to ineffective feature extraction. Lowering the radar height facilitates better feature extraction, resulting in more accurate results.

Sequence 3 involves driving along a roadway. Despite differences in point cloud density between high and low settings, the lower setting still provides enough features for EFEAR-4D to calculate the pose. Thus, the disparity between the two settings is not significant. 
 \section{conclusion}
 In this paper, an 4D radar odometry framework, EFEAR-4D, is introduced. In the ego-velocity filtering module, we employ DBSCAN clustering and utilize the least square method to fit a sphere from Doppler velocity distribution for ego-velocity estimation. Furthermore, we apply region-wise feature extraction methods to extract stable features. P2P matching is enhanced by adding a keyframe sliding window and using Huber function. Additionally, a new dataset is collected. We conduct our experiments on two public datasets and our self-collected dataset. Our method outperforms the state-of-the-art odometry methods with vision or LiDAR in different challenging environments, making 4D radar an attractive option for robust and accurate localization. 
 
 Some findings from our experiment indicate that 4D radar exhibits weaknesses in perception in certain scenarios, including instability in point cloud density and weak ground reflection due to its low power and limited FOV. These shortcomings can be mitigated by integrating complementary sensors. Future work will explore sensor fusion with 4D radar and cameras to provide a more comprehensive odometry solution.
 
\bibliographystyle{IEEEtran}
\bibliography{bib/bibliography}    
\end{document}